\documentclass{tlp}

\usepackage{aopmath}

\usepackage{amssymb}
\usepackage{xspace}
\usepackage{url}
\usepackage{color}
\usepackage{alltt}
\usepackage{multirow}
\usepackage{graphicx} 
\usepackage{bigstrut}
\usepackage{booktabs}
\usepackage[ruled,vlined]{algorithm2e}
\usepackage{ifthen}
\usepackage{tikz}
\usetikzlibrary{arrows,fit}

\newcommand{\ONLINE}[1]{See the extended version \cite{...} of this paper.}

\newtheorem{definition}{Definition} 
\newtheorem{example}{Example} 

\newcommand{\mytt}[1]{{\small\texttt{#1}}\xspace}
\newcommand{\vars}[1]{{\bf #1}\xspace}
\newcommand{\funct}[1]{\textnormal{\textsf{#1}}}
\newcommand{\at}[1]{\ensuremath{{\bf #1}}\xspace}
\newcommand{\atoms}{\func{atoms}}
\newcommand{\body}{\func{body}}
\newcommand{\head}{\func{head}}
\newcommand{\heads}{\func{heads}}
\newcommand{\base}{\func{base}}
\newcommand{\terms}{\func{terms}}
\newcommand{\subtree}{\func{subtree}}

\newcommand{\depth}{\func{depth}}
\newcommand{\guard}{\func{guard}}
\newcommand{\sides}{\func{sides}}
\newcommand{\datalogpm}{\ensuremath{\textrm{Datalog}^\pm}\xspace}
\newcommand{\datalogex}{\ensuremath{\textsl{Datalog}^\exists}\xspace}
\newcommand{\derives}{\ensuremath{\mathtt{\ \leftarrow}\ }}
\newcommand{\datalog}{\ensuremath{\textsl{Datalog}}\xspace}
\newcommand{\datalogor}{\ensuremath{\textsl{Datalog}^{\vee}}\xspace}
\newcommand{\datalogexor}{\ensuremath{\textsl{Datalog}^{\exists,\vee}}\xspace}

\newcommand{\guardeddatalogex}{\ensuremath{\guarded\emph{-}\datalogex}\xspace}
\newcommand{\guardeddatalogexor}{\ensuremath{\guarded\emph{-}\datalogexor}\xspace}
\newcommand{\lineardatalogex}{\ensuremath{\textsl{Linear}\emph{-}\datalogex}\xspace}
\newcommand{\lineardatalogexor}{\ensuremath{\textsl{Linear}\emph{-}\datalogexor}\xspace}
\newcommand{\wguardeddatalogex}{\ensuremath{\wguarded\emph{-}\datalogex}\xspace}
\newcommand{\wguardeddatalogexor}{\ensuremath{\wguarded\emph{-}\datalogexor}\xspace}

\newcommand{\mods}{\func{mods}}
\newcommand{\ans}{\func{ans}}
\newcommand{\nodes}{\func{nodes}}
\newcommand{\arcs}{\func{arcs}}
\newcommand{\Label}{\func{label}}

\newcommand{\Root}{\func{root}}

\newcommand{\TODO}[1]{{\color{red} {\Large [TODO:} #1]}\xspace}
\newcommand{\Inst}{\func{inst}\xspace}
\newcommand{\winst}{\func{winst}\xspace}

\newcommand{\guarded}{\ensuremath{\textsl{Guarded}}\xspace}
\newcommand{\wguarded}{\ensuremath{\textsl{Weakly-Guarded}}\xspace}
\newcommand{\linear}{\ensuremath{\textsl{Linear}}\xspace}

\newcommand{\coNP}{{\footnotesize \textbf{coNP}}\xspace}

\newcommand{\ACzero}{{\footnotesize \textbf{AC}$_0$}\xspace}

\newcommand{\Logspace}{{\footnotesize \textbf{LOGSPACE}}\xspace}

\newcommand{\PTIME}{{\footnotesize \textbf{P}}\xspace}
\newcommand{\EXPTIME}{{\footnotesize \textbf{EXP}}\xspace}
\newcommand{\TwoEXPTIME}{{\footnotesize \textbf{2EXP}}\xspace}
\newcommand{\PTIMEc}{\ensuremath{\PTIME{\textrm{-complete}}}\xspace}
\newcommand{\EXPTIMEc}{\ensuremath{\EXPTIME{\textrm{-complete}}}\xspace}
\newcommand{\coNPc}{\ensuremath{\coNP{\textrm{-complete}}}\xspace}
\newcommand{\EXPTIMEh}{\ensuremath{\EXPTIME{\textrm{-hard}}}\xspace}
\newcommand{\coNPh}{\ensuremath{\coNP{\textrm{-hard}}}\xspace}

\newcommand{\cmpitem}{\noindent\ensuremath{\blacktriangleright}\xspace}

\newcommand{\nop}[1]{}

\def\squareforqed{\hbox{\rlap{$\sqcap$}$\sqcup$}}
\def\qed{\ifmmode\squareforqed\else{\unskip\nobreak\hfil
\penalty50\hskip1em\null\nobreak\hfil\squareforqed
\parfillskip=0pt\finalhyphendemerits=0\endgraf}\fi}

\newcommand{\cev}[1]{\ensuremath{\overleftarrow{\;#1}}}
\newcommand{\CEV}[1]{\ensuremath{\overleftarrow{\,#1}}}

\renewcommand{\qed}{\mathproofbox}

\newcommand{\citeText}[1]{\citeANP{#1} \citeNN{#1}}
\newcommand{\beforeProof}{}

\begin{document}

\bibliographystyle{acmtrans}

\long\def\comment#1{}

\title[\datalogexor: Semantics, Decidability, and Complexity Issues]
{Disjunctive Datalog with Existential Quantifiers:\\
 Semantics, Decidability, and Complexity Issues}

\author[M. Alviano, W. Faber, N. Leone and M. Manna]
{MARIO ALVIANO, WOLFGANG FABER, NICOLA LEONE, MARCO MANNA\thanks{Marco Manna's work was supported by the European Commission through the European Social Fund and by Calabria Region.}\\
Department of Mathematics, University of Calabria, Italy\\
\email{\{alviano,faber,leone,manna\}@mat.unical.it}}

\pagerange{\pageref{firstpage}--\pageref{lastpage}}
\volume{\textbf{10} (3):}
\jdate{March 2012}
\setcounter{page}{1}
\pubyear{2012}

\maketitle

\label{firstpage}

\begin{abstract}
\datalog is one of the best-known rule-based languages, and extensions
of it are used in a wide context of applications. An important
\datalog extension is Disjunctive \datalog, which significantly increases
the expressivity of the basic language. Disjunctive \datalog is
useful in a wide range of applications, ranging from Databases (e.g.,
Data Integration) to Artificial Intelligence (e.g., diagnosis and
planning under incomplete knowledge).
However, in recent years an important shortcoming of \datalog-based
languages became evident, e.g.\ in the context of data-integration
(consistent query-answering, ontology-based data access) and
Semantic Web applications: The language does not permit any generation of
and reasoning with unnamed individuals in an obvious way. In general, it is weak
in supporting many cases of existential quantification. 
To overcome this problem, \datalogex has recently been proposed, which
extends traditional \datalog by existential quantification in rule
heads. In this work, we propose a natural extension of Disjunctive \datalog and
\datalogex, called \datalogexor, which allows both disjunctions
and existential quantification in rule heads and is therefore an
attractive language for knowledge representation and reasoning,
especially in domains where ontology-based reasoning is needed.
We formally define syntax and semantics of the
language \datalogexor, and provide a notion of instantiation, which we prove to be
adequate for \datalogexor. A main issue of \datalogex and hence also
of \datalogexor is that decidability is no longer guaranteed for
typical reasoning tasks. In order to address this issue, we identify
many decidable fragments of the language, which extend, in a natural way,
analog classes defined in the non-disjunctive case.
Moreover, we carry out an in-depth complexity analysis, deriving interesting
results which range from 
Logarithmic Space to Exponential Time. 

\medskip

\noindent \emph{To appear in Theory and Practice of Logic Programming.}
\end{abstract}
\begin{keywords}
Datalog, Non-monotonic Reasoning, Decidability, Complexity
\end{keywords}

\section{Introduction}
\datalog has its origins as a query language in Database Systems, but
the language, and in particular its extensions, have well gone beyond
this original scope, and are now used in a variety of applications,
see for example \cite{datalogreloaded}. \datalogor
\cite{EiterGottlobMannilaTODS1997}, an extension of \datalog in which
rule heads may be disjunctions of atoms, proved to be especially
rewarding in the context of AI, as it allows for the representation of
concepts like incomplete knowledge and nondeterministic effects in a
simple and intuitive way. Examples for the use of \datalogor span from
planning \cite{EiterFaberLeonePfeiferPolleresTOCL2004}, to data-integration \cite{LeoneSIGMOD2005},
to reasoning with ontologies \cite{kaon2}.

Concerning ontologies, we observe that the field of ontology-based
Query Answering (QA) is thriving in data and knowledge management
\cite{CalvaneseDeGiacomoLemboLenzeriniRosatiJAR07,CaliGottlobLukasiewiczPODS09,KolliaGlimmHorrocksDL11,CaliGottlobPierisAAAI11},
and companies such as Oracle are adding ontological reasoning modules
on top of their existing software. In this context, queries are not
merely evaluated on an extensional relational database $D$, but
against a logical theory combining 
$D$ with an \emph{ontological theory} $\Sigma$.
More specifically, $\Sigma$ describes rules and constraints
for inferring intensional knowledge from the 
data stored in $D$ \cite{JohnsonKlugJCSS1984}.
Thus, for a conjunctive query (CQ) $q$, it is not only checked whether $D$ entails $q$,
but rather whether $D\cup \Sigma$ does.

A key issue in ontology-based QA is the design of the language
used for specifying the ontological theory $\Sigma$.
To this end, \datalogpm, a family of extensions of \datalog
proposed by \citeText{CaliGottlobLukasiewiczPODS09}
for tractable QA over ontologies, has recently gained increasing interest \cite{MugnierRR2011}.
This family generalizes well-known ontology specification languages,
and is mainly based on \datalogex, an extension of \datalog
that allows existentially quantified variables in rule heads.

In this paper we propose an extension of \datalog that allows for both
disjunctions and existentially quantified variables in rule heads,
called \datalogexor.
This language is highly expressive and enables easy and powerful
knowledge-modeling, combining the ability of disjunction to deal with incomplete
information, with the power of existential quantifiers to generate
unnamed individuals and to deal with them.
For example, consider a scenario where each animal is either a carnivore or a
herbivore, and any carnivore preys at least one other animal. This knowledge
can be modeled by the following \datalogexor rules (on the left-hand side) or in
equivalent ontological terms (on the right-hand side):
\begin{alltt}\small
 carnivore(X) v herbivore(X) \(\leftarrow\) animal(X)    Animal \(\sqsubseteq\) Carnivore \(\sqcup\) Herbivore
 \(\exists\)Y preys(X,Y) \(\leftarrow\) carnivore(X)               Carnivore \(\sqsubseteq\) \(\exists\)preys.\(\top\)
 animal(Y) \(\leftarrow\) preys(X,Y)                     \(\exists\)preys\(\sp{\minus1}\).\(\top\) \(\sqsubseteq\) Animal
\end{alltt}\normalsize
In general, \datalogexor allows to naturally encode advanced ontology
properties such as role transitivity, role hierarchy, role inverse, concept
products and union of concepts.
We define the syntax of the language and provide
a formal semantics for QA over \datalogexor programs.
Since QA over \datalogexor is undecidable in the general case
(as it is undecidable already on its subclass \datalogex),
we identify a number of \datalogexor fragments where QA is decidable,
lifting to the disjunctive case several decidability results proved by
\citeText{CaliGottlobLukasiewiczPODS09}.
Moreover, we 
analyze the complexity of QA in \datalogexor by
varying different parameters.
\nop{
In this paper, we pursue the same goal
for \datalogexor. The most important difference between \datalogexor
and \datalogex is that rules are no longer deterministic, so while for
\datalogex rules can be considered as producing ground atoms, in our
work we consider \datalogexor rules as producing ground rules. Given
this abstraction, we can lift several decidability results of
\citeANP{CaliGottlobLukasiewiczPODS09}
\citeNN{CaliGottlobLukasiewiczPODS09}. To do so, we define the
central notion of instantiation, which is a set of ground rules which
allows for equivalent query answering as a given \datalogexor program
and its input facts. While this instantiation may be infinite in
general, we can show that it is finite for certain classes of programs
by specifying a bound for its creation mechanism, thus also providing
complexity results and a means for actual implementation. The main
decidable subclass that we identify is guarded \datalogexor, and we
also consider the more general class weakly guarded \datalogexor.
\TODO{Secondo me dovremmo dire qualcosa del tipo
\emph{Some of our decidability and complexity results were made possible by exploiting
notable results recently achieved by \citeANP{BaranyGottlobOttoLICS2010} \citeNN{BaranyGottlobOttoLICS2010}
on the guarded fragment of first-order logic.}}
END NOP
}
More specifically, our main contributions are the following:

\cmpitem We define the novel language \datalogexor, extending both \datalogex and \datalogor,
and provide a formal definition for QA over this language.
We also specify the notion of universal model set, which generalizes the concept of universal model
to the disjunctive case. A universal model set allows for answering any query.


\cmpitem We define the new concept of instantiation $\Inst(P)$ of a \datalogexor program $P$,
and show that it is adequate for QA.
The finiteness of $\Inst(P)$ is a sufficient condition to ensure the decidability of QA
over $P$, since one can compute a finite model set of $P$ from $\Inst(P)$ in this case.
We design a procedure for computing $\Inst(P)$ and prove that it generalizes the oblivious chase
procedure introduced by \citeText{MaierMendelzonSagivACMTODS1979} and
\citeText{JohnsonKlugJCSS1984}.

\cmpitem We define the classes of guarded, linear, and weakly guarded \datalogexor programs.
We show that: (i) they extend the corresponding classes of \datalogex programs,
(ii) checking membership in these classes is doable in polynomial time, and
(iii) QA is decidable in these classes.

\cmpitem We carry out a 
complexity analysis to determine the data complexity of QA
in all cases that are obtained by varying the following three parameters:
(i) the query (atomic, conjunctive, or acyclic),
(ii) the class of the underlying \datalogexor program
(guarded, linear, weakly guarded, monadic-linear, or multi-linear),
(iii) the allowed \datalog extension (disjunction, existential variables, or both).

To the best of our knowledge, this is the first paper proposing a dedicated extension
of Disjunctive \datalog with existential quantifiers, and analyzing its decidability
and complexity.
There have been some proposals (for example, \citeNP{FerrarisLeeLifschitzAIJ2011}) for interpreting arbitrary first-order formulas under the stable model semantics, which are more general than our approach, but have a rather different motivation and in particular do not address decidability issues.
However, in the literature there are many studies concerning the decidability of
(non-disjunctive) \datalogex fragments.
The decidable subclasses of  \datalogex rely on four main syntactic paradigms, called
\emph{guardedness} \cite{CaliGottlobKiferKR2008},
\emph{weak-acyclicity} \cite{FaginKolaitisMillerPopaTCS2005},
\emph{stickiness} \cite{CaliGottlobPierisPVLDB10}, and
\emph{shyness} \cite{LeoneKR2012}.
%
%
The guardedness paradigm will be discussed in depth in this paper and extended to the disjunctive
case.
Weak-acyclicity has originally been introduced in the context of data exchange,
where programs are required to have finite universal models \cite{FaginKolaitisMillerPopaTCS2005}.
%
Further extensions have also been proposed in this context
\cite{DeutschNashRemmelPODS2008,MarnettePODS09,MeierSchmidtLausenPVLDB09,GrecoSpezzanoTrubitsynaPVLDB11}.
Sticky \datalogex programs, defined more recently, have a low QA complexity
and can express the well-known {\em inclusion dependencies},
but, since they are FO-rewritable, they have limited expressive power.
Several generalizations of stickiness have been defined by
\citeText{CaliGottlobPierisRR10}.
For example, the \textsl{Sticky-Join} class preserves
the benign sticky complexity by also encompassing linear \datalogex programs.
Finally, \emph{Shy}, the newest among the syntactic \datalogex families,
offers a good balance between expressivity and complexity.
This class significantly extends both the class of \datalog and linear \datalogex programs,
while preserving the same (data and combined) complexity of QA over \datalog,
even though it includes existential quantifiers.

The results in this paper complement the above-mentioned works, and contribute to a more complete
picture of the computational aspects of QA over extensions of \datalog with existential
quantifiers, providing support for choosing the appropriate setting that fits particular needs in practical applications.

\section{The Disjunctive \datalogex Language}\label{sec:Framework}

In this section we introduce syntax and semantics of \datalogexor
programs and formally define the query answering problem.

\subsection{Preliminaries}

The following notation will be used throughout the paper.
We always denote
by $\Delta_C$, $\Delta_N$ and $\Delta_V$, countably infinite domains of \emph{terms} called \emph{constants}, \emph{nulls} and \emph{variables}, respectively;
by $\Delta$, the union of these three domains;
by $\varphi$, a null;
by \mytt{X} and \mytt{Y}, variables;
by $\vars{X}$ and $\vars{Y}$, sets of variables;
by $\Pi$ an alphabet of \emph{predicate symbols} each of which, say \mytt{p},
    has a fixed nonnegative arity; 
by $\at{a}$, $\at{b}$ and $\at{c}$, {\em atoms}
    being expressions of the form $\mytt{p}(t_1,\ldots,t_k)$,
    where $\mytt{p}$ is a predicate symbol,
    and $t_1,\ldots,t_k$ is a \emph{tuple} of terms.
For an atom $\at{a}$, we denote by $\funct{pred}(\at{a})$ the predicate symbol
of $\at{a}$.

For a formal structure $\varsigma$ containing atoms,
$\atoms(\varsigma)$ denotes the set of atoms in $\varsigma$, and
$\terms(\varsigma)$ denotes the set of terms occurring in $\atoms(\varsigma)$.
If $\vars{X}$ is the set of variables in $\varsigma$, i.e.,
$\vars{X} = \terms(\varsigma) \cap \Delta_V$, then $\varsigma$ is also
denoted by $\varsigma_{[\vars{X}]}$.
A structure $\varsigma_{[\emptyset]}$ is called \emph{ground}.
If $T \subseteq \Delta$ and $T \neq \emptyset$,
    then $\base(T)$ denotes the set of all atoms
    that can be formed with predicate symbols in $\Pi$ and terms from $T$.

\subsubsection{Mappings}
A \emph{mapping} is a function $\mu : \Delta \rightarrow \Delta$
s.t. $c \in \Delta_C$ implies $\mu(c) = c$,
and $\varphi \in \Delta_N$ implies $\mu(\varphi) \in \Delta_C \cup \Delta_N$.
Let $T$ be a subset of $\Delta$.
The application of $\mu$ to $T$, denoted by $\mu(T)$, is the set
$\{\mu(t) \mid t \in T\}$.
The restriction of $\mu$ to $T$, denoted by $\mu|_T$,
is the mapping $\mu'$ s.t. $\mu'(t) = \mu(t)$ for each $t \in T$, and
$\mu'(t) = t$ for each $t \notin T$.
In this case, we also say that $\mu$ is an \emph{extension} of $\mu'$,
denoted by $\mu \supseteq \mu'$.
For an atom $\at{a} = \mytt{p}(t_1,\ldots,t_k)$,
we denote by $\mu(\at{a})$ the atom $\mytt{p}(\mu(t_1),\ldots,\mu(t_k))$.
For a formal structure $\varsigma$ containing atoms,
we denote by $\mu(\varsigma)$ the structure obtained by replacing each
atom $\at{a}$ of $\varsigma$ with $\mu(\at{a})$.
The \emph{composition} of a mapping $\mu_1$ with a mapping $\mu_2$, denoted by
$\mu_2 \circ \mu_1$, is the mapping associating each $t \in \Delta$ to
$\mu_2(\mu_1(t))$.

Let $\varsigma_1$ and $\varsigma_2$ be two formal structures containing atoms.
A \emph{homomorphism} from $\varsigma_1$ to $\varsigma_2$ is a mapping $h$
s.t. $h(\varsigma_1)$ is a substructure of $\varsigma_2$
(for example, if $\varsigma_1$ and $\varsigma_2$ are sets of atoms,
$h(\varsigma_1) \subseteq \varsigma_2$).
An \emph{isomorphism} between $\varsigma_1$ and
$\varsigma_2$ is a bijective homomorphism $f$ from $\varsigma_1$ to
$\varsigma_2$. If such an isomorphism exists, $\varsigma_1$ and $\varsigma_2$ are isomorphic,
denoted by $\varsigma_1 \simeq \varsigma_2$.
%
%
A \emph{substitution} is a mapping $\sigma$ s.t.
$t \in \Delta_N$ implies $\sigma(t) = t$,
and $t \in \Delta_V$ implies $\sigma(t) \in \Delta_C \cup \Delta_N \cup \{t\}$.

\subsection{Programs and Queries}
A \datalogexor \emph{rule}
$r$ is a finite expression of the form:
\begin{equation}\label{datExOrRule}
\forall\vars{X} \exists \vars{Y} \ \ \at{disj}_{[\vars{X}' \cup \vars{Y}]} \derives  \at{conj}_{[\vars{X}]},
\end{equation}
where
(i) $\vars{X}$ and $\vars{Y}$ are disjoint sets of variables
(next called $\forall$-variables and $\exists$-variables, respectively);
(ii) $\vars{X}' \subseteq \vars{X}$;
(iii) $\at{disj}_{[\vars{X}' \cup \vars{Y}]}$ is a nonempty disjunction of atoms; and
(iv) $\at{conj}_{[\vars{X}]}$ is a conjunction of atoms.
Universal quantifiers are usually omitted to lighten the syntax,
while existential quantifiers are omitted only if $\vars{Y}$ is empty,
in which case $r$ coincides with a standard \datalogor rule.
The sets $\atoms(\at{disj}_{[\vars{X}' \cup \vars{Y}]})$ and
$\atoms(\at{conj}_{[\vars{X}]})$ are denoted by $\head(r)$ and $\body(r)$,
respectively.
If $\body(r) = \emptyset$ and $|\head(r)| = 1$, then $r$ is usually referred to
as a \emph{fact}.
In particular, $r$ is called \emph{existential} or \emph{ground} fact
according to whether $r$ contains some $\exists$-variable or not, respectively.

A \datalogexor program $P$ is a set of \datalogexor rules.
W.l.o.g., we assume that rules in $P$ do not share any variable.
We denote $\bigcup_{r \in P} \head(r)$ by $\heads(P)$.
%

A \emph{conjunctive query} (CQ) $q$, also denoted by $q(\vars{X})$, is of the form:
\begin{equation}\label{conjQuery}
    \exists\vars{Y} \ \at{conj}_{[\vars{X} \cup \vars{Y}]},
\end{equation}
where $\vars{X}$ and $\vars{Y}$ are disjoint sets of variables, and
$\at{conj}_{[\vars{X} \cup \vars{Y}]}$ is a conjunction of atoms
from $\base(\vars{X} \cup \vars{Y} \cup \Delta_C)$.
Variables in $\vars{X}$ are called \emph{free variables}.
Query $q$ is called \emph{acyclic} (ACQ, for short) if its associated hypergraph is acyclic \cite{ChekuriRajaramanTCS2000}
or, equivalently, if it has hypertree-width 1 \cite{GottlobLeoneScarcelloPODS1999}.
A \emph{Boolean CQ} (BCQ) is a query of the form (\ref{conjQuery})
s.t. $\vars{X}$ is empty.
An \emph{atomic query} is a CQ of the form (\ref{conjQuery})
s.t.\ $\at{conj}_{[\vars{X} \cup \vars{Y}]}$ consists of just one atom.

\subsection{Semantics}\label{sec:QA}



Let $M \subseteq \base(\Delta_C \cup \Delta_N)$.
$M$ is a \emph{model} of a rule
$r$ of the form (\ref{datExOrRule}), denoted by $M \models r$,
if for each substitution $\sigma$ s.t. $\sigma(\body(r)) \subseteq M$,
there is a substitution $\sigma' \supseteq \sigma|_{\vars{X}}$
s.t. $\sigma'(\head(r)) \cap M \neq \emptyset$.
$M$ is a model of a $\datalogexor$ program $P$, denoted by $M \models P$,
if $M \models r$ for each $r \in P$.
Let $\mods(P)$ denote the set of all the models of $P$.
Two programs $P,P'$ are called FO-equivalent if $\mods(P) = \mods(P')$.

A BCQ $q$ is \emph{true} w.r.t.\ a model $M$, denoted by $M \models q$,
if there is a substitution $\sigma$ s.t. $\sigma(\atoms(q)) \subseteq M$.
For a set of models $\mathcal{M}$, $q$ is true w.r.t.\ $\mathcal{M}$,
denoted by $\mathcal{M} \models q$, if $M \models q$ for each $M \in \mathcal{M}$.
For a program $P$, $q$ is true w.r.t.\ $P$, denoted by $P \models q$,
if $\mods(P) \models q$.

The answer of a CQ $q(\vars{X})$ w.r.t.\ a set of models $\mathcal{M}$,
denoted by $\ans(q,\mathcal{M})$, is the set of substitutions $\sigma|_{\vars{X}}$
s.t. $M \models \sigma|_{\vars{X}}(q)$ for each $M \in \mathcal{M}$.
The answer of $q(\vars{X})$ w.r.t.\ a program $P$,
denoted by $\ans_P(q)$, is the set $\ans(q,\mods(P))$.
Note that for a BCQ $q$, either $\ans_P(q) = \emptyset$ (if $P \not\models q$)
or $\ans_P(q) = \{\sigma|_\emptyset\}$ (if $P \models q$; $\sigma|_\emptyset$
is the identity mapping).
The same consideration also applies to $\ans(q,\mathcal{M})$.

\subsection{The Query Answering Problem}
Let $\mathcal{C}$ be a class of $\datalogexor$ programs whose terms belong
to $\Delta_C \cup \Delta_V$.
In this paper we call \emph{query answering} (QA) over $\mathcal{C}$ the
following decision problem:
Given a program $P \in \mathcal{C}$ and a BCQ $q$,
determine whether $P \models q$ holds.
In the following we will call class $\mathcal{C}$ QA-decidable if QA over
$\mathcal{C}$ is decidable.

We observe that computing $\ans_P(q)$ for a CQ $q(\vars{X})$ of the form (\ref{conjQuery})
is Turing-reducible to QA as defined above. In fact, $\ans_P(q)$
is defined as the set of substitutions $\sigma|_{\vars{X}}$ s.t.\ the BCQ
$\sigma|_{\vars{X}}(q)$ is true w.r.t.\ $P$.
Since $\sigma|_{\vars{X}} \in \ans_P(q)$ implies
$\sigma|_{\vars{X}}(\Delta_V) \subseteq \terms(P) \cap \Delta_C$,
only finitely many substitutions have to be considered.

\section{Universal Model Sets for \datalogexor Programs}\label{sec:instantiation}


In this section we generalize the notion of \emph{universal model}
widely used in the context of QA over \datalogex programs.
%
%
Intuitively, a universal model $M$ of a \datalogex program $P$ is such that
each model of $P$ is homomorphic to a subset of $M$.

\begin{definition}\label{def:universal_model_set}
Let $P \in \datalogexor$.
A set $\mathcal{M} \subseteq \mods(P)$ is a \emph{universal model set} for $P$ if
for each $M \in \mods(P)$ there is $M' \in \mathcal{M}$ and a homomorphism $h$ s.t.
$h(M') \subseteq M$. \qed
\end{definition}

Universal model sets are sufficient for QA over \datalogexor programs.

\begin{theorem}\label{thm:QA_universal}
If $\mathcal{M}$ is a universal model set for $P$,
then $P \models q$ iff $\mathcal{M} \models q$ for each BCQ $q$.
\end{theorem}

\beforeProof\begin{proof}
\noindent $(\Rightarrow)$
Immediate because $\mathcal{M} \subseteq \mods(P)$ by Definition~\ref{def:universal_model_set}.

\noindent $(\Leftarrow)$
Assume $\mathcal{M} \models q$.
Let $M$ be a model of $P$. We have to show that $M \models q$.
By Definition~\ref{def:universal_model_set}, there exist $M' \in \mathcal{M}$
and a homomorphism $h$ s.t. $h(M') \subseteq M$.
Since $\mathcal{M} \models q$ by assumption, $M' \in \mathcal{M}$ implies that
there is a substitution $\sigma$ s.t. $\sigma(\funct{atoms}(q)) \subseteq M'$.
Therefore, $h \circ \sigma(\funct{atoms}(q)) \subseteq h(M')$, and combining with
$h(M') \subseteq M$ we obtain $h \circ \sigma(\funct{atoms}(q)) \subseteq M$, i.e.,
$M \models q$.
\end{proof}


We now design a strategy for identifying a universal model set for a \datalogexor program $P$.
First, we introduce the notion of \emph{fires} of a rule $r \in P$ on a set $R$ of \datalogexor ground rules.
Next, we define an \emph{instantiation procedure} for
computing a ground program $\Inst(P)$, the models of which form a universal model set for $P$.

Let $r$ be a rule of the form (\ref{datExOrRule}),
and $R, R'$ be sets of ground rules.
A \emph{firing} substitution for $r$ w.r.t. $R$
is a substitution $\sigma$ s.t.\ $\sigma = \sigma|_{\vars{X}}$ and
$\sigma(\body(r)) \subseteq \heads(R)$.
The \emph{firing} of $r$ on $R'$ w.r.t.\ $\sigma$
yields a ground rule $\hat{\sigma}(r)$, where
$\hat{\sigma}$ is obtained by extending $\sigma|_{\vars{X}}$  as follows:
$\exists$-variables in $\vars{Y}$ are assigned
to the least $|\vars{Y}|$ nulls not occurring in $R \cup R'$.
(We assume a fixed well-ordering of $\Delta_N$ and that variables in $\vars{Y}$
are processed according to their order in $r$.)
A firing substitution for a rule $r$ is said to be \emph{spent} if it has already been fired.

\SetAlgorithmName{Procedure}{}{}
\begin{algorithm}[b]
{\small
 \LinesNumbered

 \SetKwInOut{Input}{Input}\SetKwInOut{Output}{Output}
 \Input{A \datalogexor program $P$}
 \Output{The ground program $\Inst(P)$}

  $R := \emptyset$\;
  \Repeat{$R' = \emptyset$}{
   $R' := \emptyset$\;
   \ForEach{$r \in P$ and \textnormal{\textbf{foreach}} unspent firing substitution $\sigma$ for $r$ w.r.t. $R$}{
    $R' := R' \cup \{\hat{\sigma}(r)\}$\;
   }
   $R := R \cup R'$\;
  }
  \Return{$R$;}
}
 \caption{\textsc{program-instantiation}}\label{procedure:inst}
\end{algorithm}

Procedure~1 illustrates the overall instantiation procedure.
It consists of an exhaustive series of fires in a breadth-first
(level-saturating) fashion yielding a (possibly infinite)
ground program $\Inst(P)$.

\begin{example}
Let $\varphi_1 < \varphi_2 < \cdots$ be a well-ordering of $\Delta_N$.
A run of Procedure~1 on the following program (obtained from the one given in the introduction by predicate renaming):
\begin{alltt}\small
  \(r\sb{1}:\) c(X) v h(X) \(\leftarrow\) a(X)      \(r\sb{3}:\) a(Y) \(\leftarrow\) p(X,Y)
  \(r\sb{2}:\) \(\exists\)Y p(X,Y) \(\leftarrow\) c(X)        \(r\sb{4}:\) c(lion) \(\leftarrow\)
\end{alltt}\normalsize
starts by setting $R$ and $R'$ to the empty set.
The only firing substitution w.r.t.\ $R$ is the identity substitution for
$r_4$, whose fire yields $r_4$ itself, which is then added into $R'$.
Rules in $R'$ are moved into $R$ (lines~6 and 3).
There is a new firing substitution for $r_2$, namely $\sigma_1$
s.t.\ $\sigma_1 = \sigma_1|_{\{\scriptsize \texttt{X}\}}$ and
$\sigma_1(\mytt{X}) = \mytt{lion}$.
The fire of $\sigma_1$ yields $\mytt{p(lion,}\varphi_1\mytt{)} \leftarrow \mytt{c(lion)}$,
which is added into $R'$, and then moved into $R$.
Now there is a firing substitution for $r_3$, namely
$\sigma_2$ s.t.\ $\sigma_2 = \sigma_2|_{\{\scriptsize \texttt{X},\texttt{Y}\}}$,
$\sigma_2(\mytt{X}) = \mytt{lion}$ and
$\sigma_2(\mytt{Y}) = \varphi_1$,
whose fire yields $\mytt{a(}\varphi_1\mytt{)} \leftarrow \mytt{p(lion,}\varphi_1\mytt{)}$.
After adding this rule into $R'$, and then moving it into $R$,
there is a new firing substitution for $r_1$, namely
$\sigma_3$ s.t.\ $\sigma_3 = \sigma_3|_{\{\scriptsize \texttt{X}\}}$ and
$\sigma_3(\mytt{X}) = \varphi_1$.
The fire of $\sigma_3$ yields $\mytt{c(}\varphi_1\mytt{) v h(}\varphi_1\mytt{)} \leftarrow \mytt{a(}\varphi_1\mytt{)}$,
which is added into $R'$, and then moved into $R$.
Now there is a new firing substitution for $r_2$, namely
$\sigma_4$ s.t.\ $\sigma_4 = \sigma_4|_{\{\scriptsize \texttt{X}\}}$ and
$\sigma_4(\mytt{X}) = \varphi_1$,
whose fire yields $\mytt{p(}\varphi_1\mytt{,}\varphi_2\mytt{)} \leftarrow \mytt{c(}\varphi_1\mytt{)}$.
The procedure thus go on, indefinitely.
Let $I = \{\mytt{c(lion)}, \ \mytt{p(lion,}\varphi_1\mytt{)}, \ \mytt{a(}\varphi_1\mytt{)}\}$.
Subset-minimal models of $\Inst(P)$ have the following forms:
$$
\begin{array}{clllr}
\bullet & \bigcup_{i \in [1\mbox{..}k]} & \{\mytt{c(}\varphi_i\mytt{)}, \ \mytt{p(}\varphi_i\mytt{,}\varphi_{i+1}\mytt{)}, \ \mytt{a(}\varphi_{i+1}\mytt{)}\} \ \cup \ I \ \cup \
\{\mytt{h(}\varphi_{k+1}\mytt{)}\}, & \forall k \geq 1; &  \\
\bullet & \bigcup_{i \geq 1} & \{\mytt{c(}\varphi_i\mytt{)}, \ \mytt{p(}\varphi_i\mytt{,}\varphi_{i+1}\mytt{)}, \ \mytt{a(}\varphi_{i+1}\mytt{)}\} \ \cup \ I\mbox{.} & \ & \Box
\end{array}
$$
\end{example}

\nop{
The \emph{level} of a rule in $\Inst(P)$ is inductively defined as follows:
Each rule with an empty body has level $1$.
The level of each rule constructed after a given firing
is obtained from the highest level of the
atoms in $\sigma(\body(r))$ plus one.
For each $k \geq 0$, $\Inst^k(P)$ denotes the subset of $\Inst(P)$
containing only and all the rules of level up to $k$ where,
by convention, $\Inst^0(P) = \emptyset$.
Actually, by \mbox{Procedure 1}, $\Inst^k(P)$ is precisely the set of rules which is inferred
the $k^{th}$-time that the outer for-loop is ran.
}

In order to show that $\mods(\Inst(P))$ is a universal model set for $P$,
we first point out some relationships between the models of $P$ and those of $\Inst(P)$.

\begin{lemma}\label{lem:semeq:second}
Let $P$ be a \datalogexor program and $P' = \Inst(P)$.
For each $M \in \mods(P)$ there exist $M' \in \mods(P')$
and a homomorphism $h$ s.t.:
(i) $M' \subseteq \heads(P')$;
(ii) $h(M') \subseteq M$; and
(iii) $h = h|_{\terms(P')}$.
\end{lemma}
\beforeProof\begin{proof}
Let $M \in \mods(P)$ and
$P_i = \{r_1, \ldots, r_i\}$ be the first $i$ rules in $P'$
(w.r.t. the order induced by Procedure~1).
We prove by induction that, for each $i\geq 0$, there exist
$M_i \in \mods(P_i)$ and a homomorphism $h_i$ s.t.:
$M_i \subseteq \heads(P_i)$;
$h_i(M_i) \subseteq M$; and
$h_i = h_i|_{\terms(P_i)}$.

The base case, for $i = 0$, is vacuously true by choosing $M_0 = \emptyset$ and
$h_0$ the identity mapping.
Let us assume that the claim holds for some $i \geq 0$,
and let us extend $M_i$ and $h_i$ in order to show that the claim holds for $i+1$.

Note that rule $r_{i+1}$ has been obtained by a substitution
$\hat{\sigma}$ and a rule $r \in P$ of the form (\ref{datExOrRule}).
Note also that $h_i \circ \hat{\sigma}$ is a substitution because
$h_i = h_i|_{\terms(P_i)}$ by the induction hypothesis.
If $h_i \circ \hat{\sigma}(\body(r)) \subseteq M$, there is a substitution
$\sigma' \supseteq (h_i \circ \hat{\sigma})|_{\vars{X}}$
s.t. $\sigma'(\head(r)) \cap M \neq \emptyset$
(because $M$ is a model of $P$ by assumption).
Otherwise, if $h_i \circ \hat{\sigma}(\body(r)) \nsubseteq M$,
let $\sigma' = h_i \circ \hat{\sigma}$.
Let $h_{i+1}$ be the homomorphism s.t.
$t \in \hat{\sigma}(\vars{Y})$ implies $h_{i+1}(t) = \sigma'(t)$, and
$t \notin \hat{\sigma}(\vars{Y})$ implies $h_{i+1}(t) = h_i(t)$.
Let $M_{i+1}$ be the following set of atoms:
$M_i \cup \hat{\sigma}(\{\at{a} \in \atoms(r) \mid \sigma'(\at{a}) \in M\})$.

The following properties hold by construction:
$M_{i+1} \subseteq \heads(P_{i+1})$;
$h_{i+1}(M_{i+1}) \subseteq M$; and
$h_{i+1} = h_{i+1}|_{\terms(P_{i+1})}$.
Hence, to complete the proof, we have just to prove that $M_{i+1}$ is a model
of $P_{i+1}$. In fact, this is the case because:
$r_{i+1}$ is satisfied by construction of $M_{i+1}$;
rules of $P_i$ are satisfied by $M_{i+1}$ because they are satisfied by $M_i$,
and atoms in $M_{i+1} \setminus M_i$ do not occur in $P_i$ by construction of
$M_{i+1}$.
\end{proof}

A
universal model set for $P$
can be obtained from $\mods(\Inst(P))$, which allows for answering queries on $P$
by performing the reasoning on $\Inst(P)$.

\begin{theorem}\label{thm:semeq}
Let $P$ be a \datalogexor program and $P' = \Inst(P)$.
Model set $\mathcal{M} = \{M \in \mods(P') \mid M \subseteq \heads(P')\}$ is universal for $P$.
\end{theorem}
\beforeProof\begin{proof}
By Lemma~\ref{lem:semeq:second}, for each $M \in \mods(P)$ there is
$M' \in \mathcal{M}$ and a homomorphism $h$ s.t. $h(M') \subseteq M$.
It remains to show that $\mathcal{M} \subseteq \mods(P)$, i.e.,
$M \in \mods(P')$ s.t.\ $M \subseteq \heads(P')$ implies $M \in \mods(P)$.
Let $r \in P$ and $\sigma$ be a substitution s.t. $\sigma(\body(r)) \subseteq M$, so $\sigma$ is a firing substitution for $P'$.
Let $\hat{\sigma}(r)$ be the rule of $P'$ obtained by the firing of $r$.
Thus, $\head(\hat{\sigma}(r)) \cap M \neq \emptyset$, i.e.,
$M \models \hat{\sigma}(r)$.
\end{proof}

The program produced by Procedure~1 is a generalization of the oblivious chase
procedure \cite{MaierMendelzonSagivACMTODS1979,JohnsonKlugJCSS1984}, which
associates every \datalogex program with a universal model.
In fact, the oblivious chase procedure can be obtained from Procedure~1 by
replacing line~5 with $R' := R' \cup \hat{\sigma}(\head(r))$, which is
enough for \datalogex programs.

\begin{corollary}
Let $P$ be a \datalogex program. Then, $\{\heads(\Inst(P))\}$ is universal for $P$.
\end{corollary}

\section{Extending guards-based classes to \datalogexor}\label{sec:classes}

We next define subclasses of \datalogexor
relying on a well known paradigm, called \emph{guardedness},
first introduced by \citeANP{AndrekaNemetiVanBenthem1998} \citeNN{AndrekaNemetiVanBenthem1998}
in the definition of the guarded fragment of first-order logic
and further revisited by \citeANP{CaliGottlobKiferKR2008} \citeNN{CaliGottlobKiferKR2008}
for defining \datalogex subclasses.
%
%
%
In the next section, we show that all these new classes both
depend on (easily) checkable syntactic properties, and
are QA-decidable.

\begin{definition}\label{def:guarded}
A \datalogexor rule $r$ is said to be \emph{guarded} if it is of the form:
\begin{equation}\label{eq:guarded_rule}
\forall\vars{X} \exists \vars{Y} \ \at{disj}_{[\vars{X}' \cup \vars{Y}]} \derives
     \ \at{guard}_{[\vars{X}]},\ \at{s\emph{-}conj}_{[\vars{X}'']},
\end{equation}
where
$\vars{X}'$ and $\vars{X}''$ are subsets of $\vars{X}$,
$\at{guard}_{[\vars{X}]}$ is an atom called \emph{guard} and denoted by $\guard(r)$,
$\at{s\emph{-}conj}_{[\vars{X}'']}$ is a conjunction of atoms called \emph{sides}
and denoted by $\sides(r)$.
Moreover, a guarded rule $r$ is called:
\emph{multi-linear} if each side atom could be chosen as guard;
\emph{linear} if $\sides(r) = \emptyset$;
\emph{monadic-linear} if $\sides(r) = \emptyset$ and
all head predicates are unary.
Hereafter, a \datalogexor program $P$ is called \guarded
(resp., \textsl{Multi-Linear}, \linear, \textsl{Monadic-}\linear) if
each rule $r \in P$ either is guarded (resp., multi-linear, linear, monadic-linear)
or has an empty body. \qed
\end{definition}

We now introduce the notion of affected positions of an atom,
which are the only positions where nulls might occur in the output of
Procedure~1.

\begin{definition}\label{def:affected}
Let $P$ be a \datalogexor program,
$\at{a}$ be an atom,
and $\mytt{X}$ a variable occurring in $\at{a}$ at position $i$.
Position $i$ of $\at{a}$ is (inductively) marked as \emph{affected} w.r.t. $P$
if there is a rule $r \in P$ with an atom $\at{b} \in \head(r)$
s.t.
$\funct{pred}(\at{b}) = \funct{pred}(\at{a})$ and
$\mytt{X}$ is either
an $\exists$-variable,
or a $\forall$-variable s.t. $\mytt{X}$ occurs in $\body(r)$ in affected positions only.
A variable \mytt{X} occurring in the body of a rule is
\emph{unaffected} if it is not affected.\qed
\end{definition}

The above definition is now used
to define the class of weakly-guarded programs.

\begin{definition}\label{def:wguarded}
Let $P$ be a \datalogexor program, and $r \in P$ be a rule of the form:
\begin{equation}\label{eq:wguarded_rule}
\forall \vars{X} \exists \vars{Y} \ \at{disj}_{[\vars{X}' \cup \vars{Y}]} \derives
    \ \at{wguard}_{[\vars{X}'']}, \at{s\emph{-}conj}_{[\vars{X}''']},
\end{equation}
where $\vars{X}' \subseteq \vars{X} = \vars{X}'' \cup \vars{X}''' $.
Rule $r$ is said to be \emph{weakly-guarded} w.r.t. $P$, if each variable in
$\vars{X}''' \setminus \vars{X}''$ is unaffected in $r$.
Here, $\guard(r)$ and $\sides(r)$
still denote the (weak) guard and the side atoms of $r$, respectively.
In the following, \wguardeddatalogexor will denote
the set of \datalogexor programs where each rule either is weakly-guarded or has an empty body.\qed
\end{definition}

The new \datalogexor subclasses introduced in this section generalize
important fragments of \guardeddatalogex already analyzed in the literature.
(Note \ that \wguardeddatalogexor generalized \wguardeddatalogex because
\ for disjunction-free programs
Definition~\ref{def:affected} coincides with the
the notion of affected position introduced by
\citeNP{CaliGottlobKiferKR2008}.)

\begin{proposition}
Definitions \ref{def:guarded} and \ref{def:wguarded} generalize
the classes \guardeddatalogex, \lineardatalogex, and \wguardeddatalogex
defined by \citeText{CaliGottlobKiferKR2008}.
\end{proposition}

We now pinpoint the complexity of recognizing programs in these classes.

\begin{theorem}
Checking whether a program belongs to \guardeddatalogexor,
\lineardatalogexor, or \wguardeddatalogexor is decidable,
and doable in polynomial-time.
\end{theorem}

\beforeProof\begin{proof}
Checking whether a program is guarded (resp., linear or multi-linear) is doable in
linear time by inspection of the rule bodies.
Concerning a \wguardeddatalogexor program $P$, we observe that
Definition~\ref{def:affected} introduces a monotone operator for determining
affected positions, and the number of such positions is linear in the size of $P$.
Hence, all affected positions in $P$ can be determined
in quadratic time.
\end{proof}

\section{Decidability Results}

We now show that all classes introduced in the previous section are
QA-decidable. In particular, we use results recently established by
\citeText{BaranyGottlobOttoLICS2010}
on the \emph{guarded fragment of first-order logic}
\cite{AndrekaNemetiVanBenthem1998,Gradel1999},
here denoted by \textsl{Guarded-FOL} and inductively defined as follows:
(i) $\base(\Delta_C \cup \Delta_V) \subset \textsl{Guarded-FOL}$;
(ii) if $\psi_1, \psi_2 \in \textsl{Guarded-FOL}$, then
    $\neg \psi_1$, $\psi_1 \vee \psi_2$, $\psi_1 \wedge \psi_2$, and $\psi_1 \leftarrow \psi_2$ also belong to \textsl{Guarded-FOL}; and
(iii) if $\at{a}_{[\vars{X}\cup\vars{Y}]} \in \base(\Delta_C \cup \Delta_V)$, $\psi(\vars{X}'\cup\vars{Y}') \in \textsl{Guarded-FOL}$,
    and the (free) variables of $\at{a}$ include all the free variables $\vars{X}'\cup\vars{Y}'$ of $\psi$,
    then $\exists \vars{Y} (\at{a}_{[\vars{X}\cup\vars{Y}]} \wedge \psi(\vars{X}'\cup\vars{Y}'))$ and
    $\forall \vars{X} (\psi(\vars{X}'\cup\vars{Y}') \derives \at{a}_{[\vars{X}\cup\vars{Y}]})$ are also in \textsl{Guarded-FOL}.

Any \guardeddatalogexor program can be viewed as a \textsl{Guarded-FOL} formula.

\begin{proposition}\label{prop:GuardedIsGFO}
There is a logarithmic space transducer associating each \guardeddatalogexor
program with a FO-equivalent \textsl{Guarded-FOL} formula.
\end{proposition}

\beforeProof\begin{proof}
For a guarded \datalogexor rule $r$ of the form (\ref{eq:guarded_rule}),
let $\at{h}^i_{[\vars{X}'_i \cup \vars{Y}_i]}$ be the $i$-th atom in $\at{disj}_{[\vars{X}' \cup \vars{Y}]}$,
with \mbox{$i \in [1\emph{..}k]$,} $\vars{X}'_i \subseteq \vars{X}'$, and $\vars{Y}_i \subseteq \vars{Y}$.
Rule $r$ is translated into the following FO-equivalent formula:
\[
    \forall\vars{X} ( \exists \vars{Y}_1 \at{h}^1_{[\vars{X}'_1 \cup \vars{Y}_1]} \vee \cdots \vee \exists \vars{Y}_k \at{h}^k_{[\vars{X}'_k \cup \vars{Y}_k]} \vee \neg \at{s\emph{-}conj}_{[\vars{X}'']} \derives  \at{guard}_{[\vars{X}]}).
\]
The whole disjunction is an expression $\psi(\vars{X}' \cup \vars{X}'')$ in \textsl{Guarded-FOL} because
each $\exists \vars{Y}_i \at{h}^i_{[\vars{X}'_i \cup \vars{Y}_i]}$ is equivalent to $\exists \vars{Y}_i (\at{h}^i_{[\vars{X}'_i \cup \vars{Y}_i]} \wedge \at{h}^i_{[\vars{X}'_i \cup \vars{Y}_i]}) \in \textsl{Guarded-FOL}$, and since
$\neg \at{s\emph{-}conj}_{[\vars{X}'']}$ trivially belongs to $\textsl{Guarded-FOL}$.
Moreover, the expression $\forall \vars{X} (\psi(\vars{X}' \cup \vars{X}'') \derives \at{guard}_{[\vars{X}]})$ is in $\textsl{Guarded-FOL}$
since $\vars{X}' \cup \vars{X}'' \subseteq \vars{X}$.
Finally, a similar construction applies to rules having empty bodies.
\end{proof}

QA-decidability of \guardeddatalogexor and its subclasses can now be established.

\begin{theorem}\label{thm:GuardedDecidable}
Conjunctive QA is decidable under \guarded, \textsl{Multi-}\linear and \linear
\datalogexor.
\end{theorem}

\beforeProof\begin{proof}
The result follows from Proposition \ref{prop:GuardedIsGFO} and
from the fact that conjunctive QA
is decidable under \textsl{Guarded-FOL} \cite{BaranyGottlobOttoLICS2010}.
\end{proof}


In order to prove that $\wguarded$-\datalogexor is QA-decidable as well,
we first introduce the notion of \emph{weak instantiation}.

\begin{definition}\label{def:winst}
Let $P \in \wguarded$-\datalogexor.
For each $r \in P$, let $\winst(r)$ denote the set of partially ground rules
associated to $r$ and consisting of the set $\{r\}$ or of the set
$\{\sigma(r) \mid \sigma \ \textrm{is a substitution from} \ \vars{X}''' \setminus \vars{X}'' \ \textrm{to} \ \terms(P) \cap \Delta_C\}$
according to whether rule $r$ has an empty body or is of the form (\ref{eq:wguarded_rule}), respectively.
The weak instantiation of $P$, denoted by $\winst(P)$, is defined
as the union of $\winst(r)$ for each $r \in P$.\qed
\end{definition}

The above definition transforms any $\wguarded$-\datalogexor program
into a FO-equivalent \guardeddatalogexor program.

\begin{lemma}\label{lem:decidWGard}
Let $P$ be a $\wguarded$-\datalogexor program and $P' = \winst(P)$.
Then, both $P' \in \guardeddatalogexor$ and $\Inst(P) \simeq \Inst(P')$ hold.
\end{lemma}

\beforeProof\begin{proof}
Assume that Procedure~1 builds isomorphic sets of rules for $P$ and $P'$ up to
a given iteration of the repeat-until loop. We shall show that this isomorphism
can be extended to the succeeding iteration.
For each firing substitution $\sigma$ for a rule $r \in P$, there are
$\sigma_1, \sigma_2$ s.t.\ $\sigma = \sigma_2 \circ \sigma_1$, where
$\sigma_1$ is a substitution from $\vars{X}''' \setminus \vars{X}''$ to
$\terms(P) \cap \Delta_C$.
Let $r' = \sigma_1(r)$.
Therefore, $r' \in P'$ and $\sigma_2$ is a firing substitution for $r'$.
Consider now the other direction.
Let $\sigma'$ be a firing substitution for $r' \in P'$.
Let $r' = \sigma(r)$, where $r \in P$ and $\sigma$ is a substitution from
$\vars{X}''' \setminus \vars{X}''$ to $\terms(P) \cap \Delta_C$.
Therefore, $\sigma' \circ \sigma$ is a firing substitution for $r$.
The isomorphism can thus be extended by opportunely mapping new nulls.
\end{proof}

We can thus conclude that \wguardeddatalogexor is QA-decidable.

\begin{theorem}\label{thm:decidWGard}
Conjunctive QA is decidable under \wguardeddatalogexor.
\end{theorem}

\beforeProof\begin{proof}
The statement directly follows from Lemma \ref{lem:decidWGard} and Theorem \ref{thm:GuardedDecidable}.
\end{proof}

\section{Complexity Analysis}

In this section we study data complexity of QA
under different classes of \datalogexor and queries.
As usual in this setting, we assume that a \datalogexor program $P$ is paired
with a (finite) database $D \subset \base(\Delta_C)$.
The set of ground facts $\{\at{a} \derives \mid \at{a} \in D\}$
is denoted by $\CEV{D}$.
Similarly, $\cev{\at{a}}$ denotes the singleton $\{\at{a} \derives\}$ for some atom $\at{a} \in D$.
Finally, whenever $P$ contains a rule $r$ of the form $\at{disj} \derives$ (even if $|\at{disj}| = 1$),
we replace it in $P$ by $\at{disj} \derives \mytt{edb}$ and we add to $D$ the extra
(propositional) atom \mytt{edb} of arity zero.
Hereafter, we assume $D = \{\at{a}_1,\ldots,\at{a}_n\}$.

\subsection{\guardeddatalogexor}\label{sec:complexGuard}

We start by providing an upper bound for QA over \guardeddatalogexor.

\begin{theorem}\label{thm:MembershipFromLICS}
Data complexity of QA over \guardeddatalogexor programs is in \coNP.
\end{theorem}

\beforeProof\begin{proof}
From statement 5 of Theorem 19 in \citeText{BaranyGottlobOttoLICS2010},
data complexity of deciding whether a CQ is true w.r.t.\ a \textsl{Guarded-FOL}
formula is in \coNP.
The claim therefore follows from Proposition~\ref{prop:GuardedIsGFO}.
\end{proof}

We now pinpoint the complexity of QA over \guardeddatalogexor.

\begin{theorem}\label{thm:coNPcompl}
Data complexity of QA over \guardeddatalogexor programs is \coNPc in general,
and it is \coNPh already in the following cases:
\begin{enumerate}
  \item A \textsl{Monadic-Linear-}\datalogor program
    under an acyclic CQ.

  \item A \textsl{Multi-Linear-}\datalogor program under an atomic query.

\end{enumerate}
\end{theorem}

\beforeProof\begin{proof}
\noindent $(1)$
QA is \coNP-hard already in the following setting:
a database and an acyclic CQ
involving only unary and binary atoms, and
a single (nonrecursive) \textsl{Monadic-Linear-}\datalogor rule containing two head atoms.
This result follows from Theorem 6.4 (and its proof) of
\citeText{CalvaneseDeGiacomoLemboLenzerini2009}:
%
Let $\phi$ be a 2+2-CNF formula,
namely a CNF formula where each clause has exactly two positive and two negative literals.
Let $D$ be a database containing
an atom $\mytt{lit(}x\mytt{)}$ for each propositional variable $x$, and
atoms
$\mytt{p}_1\mytt{(}c\mytt{,}x_1\mytt{)}$,
$\mytt{p}_2\mytt{(}c\mytt{,}x_2\mytt{)}$,
$\mytt{n}_1\mytt{(}c\mytt{,}x_3\mytt{)}$,
$\mytt{n}_2\mytt{(}c\mytt{,}x_4\mytt{)}$
for each clause $x_1 \vee x_2 \vee \neg~x_3 \vee \neg~x_4$
having $c$ as identifier.
Let $P$ be a \textsl{Monadic-Linear-}\datalogor program consisting of the following rule:
$\mytt{t(X) v f(X)} \leftarrow \mytt{lit(X)}$, and
$q$ be the following acyclic CQ:
$\exists \ \mytt{C,P}_1\mytt{,P}_2\mytt{,N}_1\mytt{,N}_2$ \
$\mytt{p}_1\mytt{(C},\mytt{P}_1\mytt{)}\mytt{,}\mytt{f(P}_1\mytt{),p}_2\mytt{(C,P}_2\mytt{),f(P}_2\mytt{),n}_1\mytt{(C,N}_1\mytt{),t(N}_1\mytt{),n}_2\mytt{(C,N}_2\mytt{),t(N}_2\mytt{)}$.
Hence, $\phi$ is unsatisfiable if and only if $P \cup \CEV{D} \models q$.
\medskip

\noindent $(2)$
The \coNPc problem \textsc{3-unsat} can be encoded by means of an atomic query
$\mytt{wrongAssignment}$ over the following \textsl{Multi-Linear-}\datalogor
program $P$:
\begin{alltt}\small
  sel(L\(\sb{1}\),N\(\sb{1}\)) v sel(L\(\sb{2}\),N\(\sb{2}\)) v sel(L\(\sb{3}\),N\(\sb{3}\)) \(\leftarrow\) clause(L\(\sb{1}\),L\(\sb{2}\),L\(\sb{3}\),N\(\sb{1}\),N\(\sb{2}\),N\(\sb{3}\)).
  wrongAssignment \(\leftarrow\) sel(L,N), sel(N,L).
\end{alltt}\normalsize
As far as database $D$ is concerned, each clause $\ell_1 \vee \ell_2 \vee \ell_3$
of a given 3-CNF formula $\phi$ is
encoded in $D$ by the ground atom
$\mytt{clause}(``\ell_1",``\ell_2",``\ell_3",n(\ell_1),n(\ell_2),n(\ell_3))$,
where $n(\ell) = ``\neg x"$ if $\ell$ is a positive propositional variable $x$,
and $n(\ell) = ``x"$ if $\ell$ is a negative propositional variable $\neg x$.
If there is a satisfying assignment for $\phi$, then there is a model of $P \cup \CEV{D}$
not containing $\mytt{wrongAssignment}$.
%
\end{proof}

\subsection{\wguardeddatalogexor}\label{sec:complexWGuard}

%

As in the disjunction-free case, the complexity of QA over \wguardeddatalogexor
is harder than QA over \guardeddatalogexor.

\begin{theorem}\label{thm:expcompl}
Data complexity of QA over \wguardeddatalogexor is \EXPTIMEc in general,
and it is \EXPTIMEh already for atomic queries over \wguardeddatalogex.
\end{theorem}

\beforeProof\begin{proof}
Hardness comes from the {\bf \small EXP}-hardness of \wguarded-\datalogex \cite{CaliGottlobKiferKR2008}.
As for the membership, let $P$ be a $\wguarded$-$\datalogexor$ program
and $P' = \winst(P \cup \CEV{D})$ be the \guarded-\datalogexor program built according to Definition \ref{def:winst}.
By Lemma~\ref{lem:decidWGard}, $P\cup \CEV{D} \models q$ iff $P' \models q$.
Moreover, let $k$ be the maximum number of unguarded (thus unaffected) variables appearing in some rule of $P$,
$\gamma$ be the number of constants occurring in $P$, and
$w$ be the maximum arity over all predicate symbols in $P \cup \CEV{D}$.
We point out that \mbox{$|P'| \leq  |D| + |P|\cdot (w\cdot|D| + \gamma)^k$}.
Hence, in data complexity, the size of $P'$ is polynomial in the cardinality of $D$.
\citeANP{BaranyGottlobOttoLICS2010} \citeNN{BaranyGottlobOttoLICS2010} have
shown that QA over a \textsl{Guarded-FOL} formula
is in \TwoEXPTIME in the general case.
However, this double exponential dependence is only in terms
of $q$ and $w$. If $P$ and $q$ are considered fixed, then
the complexity is simply exponential in the size of $P'$.
Moreover, since $P'$ can be translated in logarithmic space into a
FO-equivalent $\textsl{Guarded-FOL}$ formula by
Proposition~\ref{prop:GuardedIsGFO}, then
we have an \EXPTIME (w.r.t. the cardinality of $D$) algorithm
deciding whether $P' \models q$.
\end{proof}

\subsection{Atomic Queries over \lineardatalogexor}\label{sec:complexAtomLin}

In the following, let $P$ be a \lineardatalogexor program and $q$ be Boolean
atomic query.
As before, $D = \{\at{a}_1,\ldots,\at{a}_n\}$ is the input database.
We first introduce a decomposition property relying on the structure of $P$.

\begin{lemma}\label{lem:decomp}
Let $C$ be the set $\mods(P \cup \cev{\at{a}_1}) \times \cdots \times \mods(P \cup \cev{\at{a}_n})$, and
$\mathcal{M}$ be $\{M_1 \cup \cdots \cup M_n  \mid \langle M_1, \ldots, M_n \rangle \in C\}$.
It holds that $\mathcal{M} = \mods(P \cup \CEV{D})$.
\end{lemma}

\beforeProof\begin{proof}
$(\subseteq)$
Let $\langle M_1, \ldots, M_n \rangle \in C$, and $M = M_1 \cup \cdots \cup M_n$.
To prove that $M$ is a model of $P \cup \CEV{D}$, we have to show that
whenever for a rule $r \in P$ there exists a substitution $\sigma$ s.t. $\sigma(\body(r)) \subseteq M$,
then $M \models \sigma(\head(r))$.
Let us fix a pair $(r,\sigma)$ s.t. $\sigma(\body(r)) \subseteq M$.
Since $P$ is linear, there is $i \in [1\emph{..}n]$ s.t. $\sigma(\body(r)) \subseteq M_i$.
But since $M_i$ is a model of $P \cup \cev{\at{a}}$, then $M_i \models \sigma(\head(r))$.
Finally, the implication holds since $M_i \subseteq M$.

\medskip
\noindent $(\supseteq)$
Let $M$ be a model of $P \cup \CEV{D}$. For each $i \in [1\emph{..}n]$, $M$ is also a model of  $P \cup \cev{\at{a}_i}$.
Consequently, the $n$-tuple $\langle M, \ldots, M \rangle$ belongs to $C$, entailing
that $M \in \mathcal{M}$.
\end{proof}

The following lemma represents a logspace Turing reduction
from the problem of evaluating $q$ over $P \cup \CEV{D}$
to the problem of evaluating $q$ over $P \cup \CEV{\at{a}}$ for some $\at{a} \in D$.

\begin{lemma}\label{lem:QAOneAtom}
$P \cup \CEV{D} \models q$ if and only if $\exists i \in [1\emph{..}n]$ s.t. $P \cup \cev{\at{a}_i} \models q$.
\end{lemma}

\beforeProof\begin{proof}
$(\Rightarrow)$
We prove the contrapositive. Let us assume that $\forall i \in [1\emph{..}n]$
$P \cup \cev{\at{a}_i} \not \models q$.
Thus, $\forall i \in [1\emph{..}n]$ there exists a model
$M_i$ s.t. $M_i \not \models q$.
Therefore, $M_1 \cup \cdots \cup M_n \not\models q$ and
by Lemma \ref{lem:decomp} we obtain $P \cup \CEV{D} \not\models q$.

\medskip
\noindent $(\Leftarrow)$
Since $\exists i \in [1\emph{..}n]$ s.t. $P \cup \cev{\at{a}}_i \models q$,
then $M \models q$ for each $M \in \mods(P \cup \cev{\at{a}}_i)$.
By Lemma \ref{lem:decomp}, $P \cup \CEV{D} \models q$.
\end{proof}

Lemma \ref{lem:QAOneAtom} allows for focussing the analysis on a single database atom, say $\at{a} \in D$.
The \emph{instantiation-tree} for $P \cup \cev{\at{a}}$
is the directed acyclic graph $T = \funct{tree}(P \cup \cev{\at{a}})$
inductively constructed as follows:
(i) the root of $T$ is a node labeled with $\cev{\at{a}}$;
(ii) for each node $m$ of $T$ and for each rule $r \in \Inst(P \cup \cev{\at{a}})$
s.t. $\body(r)$ appears in the head of the rule labeling $m$, we add a node $n$ labeled with $r$
along with an arc from $m$ to $n$.
(See Example~\ref{ex:instantiation_tree}.)
%
%
Let $\nodes(T)$ and $\arcs(T)$ denote the nodes and arcs of $T$, respectively;
$\Label(n)$ denotes the ground rule used as label for $n$;
$n \in T$ is short for $n \in \nodes(T)$;
$\funct{subtree}(n)$ is the tree below $n$;
finally, $\depth(n)$ is the \emph{depth} of $n$ in $T$,
defined as the length of the path leading from the root of $T$ to $n$.

\begin{definition}
The \emph{stem} of $P \cup \cev{\at{a}}$, denoted by $\funct{stem}(P \cup \cev{\at{a}})$,
is the maximal subtree that can be obtained starting from the root of $\funct{tree}(P \cup \cev{\at{a}})$
in such a way that each path contains no nodes labelled with rules with isomorphic bodies.
Finally, $\funct{sinst}(P \cup \cev{\at{a}})$ denotes the set
$\{\Label(n) \mid n \in \funct{stem}(P \cup \cev{\at{a}})\}$. \qed
\end{definition}

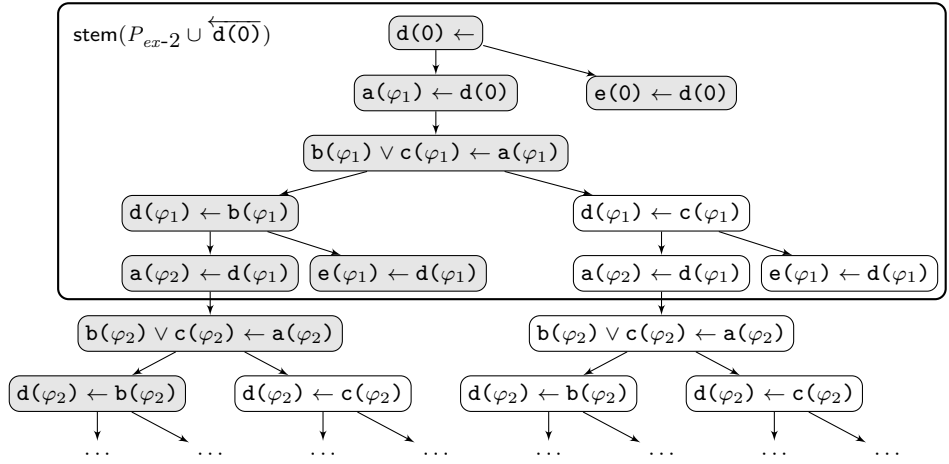
\begin{figure}[t]
\footnotesize
\tikzstyle{node} = [rectangle, draw, text centered, rounded corners]
\tikzstyle{C_0} = [rectangle, draw, fill=gray!20, text centered, rounded corners]
\tikzstyle{none} = [text centered]
\tikzstyle{line} = [draw, -latex']
\tikzstyle{bigbox} = [draw, thick, rounded corners, rectangle]

\begin{tikzpicture}
\node at (-3.5,0) [none](label) {$\funct{stem}(P_{ex\mbox{-}\ref{ex:instantiation_tree}} \cup \cev{\mytt{d(0)}})$};

\node at (0,0) [C_0](A) {$\mytt{d(0)} \leftarrow$};

\node at (0,-.8) [C_0](AA) {$\mytt{a(}\varphi_1\mytt{)} \leftarrow \mytt{d(0)}$};
\node at (3,-.8) [C_0](AB) {$\mytt{e(0)} \leftarrow \mytt{d(0)}$};

\node at (0,-1.6) [C_0](AAA) {$\mytt{b(}\varphi_1\mytt{)} \vee \mytt{c(}\varphi_1\mytt{)} \leftarrow \mytt{a(}\varphi_1\mytt{)}$};

\node at (-3,-2.4) [C_0](AAAA) {$\mytt{d(}\varphi_1\mytt{)} \leftarrow \mytt{b(}\varphi_1\mytt{)}$};
\node at (3,-2.4) [node](AAAB) {$\mytt{d(}\varphi_1\mytt{)} \leftarrow \mytt{c(}\varphi_1\mytt{)}$};

\node at (-3,-3.2) [C_0](AAAAA) {$\mytt{a(}\varphi_2\mytt{)} \leftarrow \mytt{d(}\varphi_1\mytt{)}$};
\node at (-.5,-3.2) [C_0](AAAAB) {$\mytt{e(}\varphi_1\mytt{)} \leftarrow \mytt{d(}\varphi_1\mytt{)}$};
\node at (3,-3.2) [node](AAABA) {$\mytt{a(}\varphi_2\mytt{)} \leftarrow \mytt{d(}\varphi_1\mytt{)}$};
\node at (5.5,-3.2) [node](AAABB) {$\mytt{e(}\varphi_1\mytt{)} \leftarrow \mytt{d(}\varphi_1\mytt{)}$};

\node at (-3,-4) [C_0](AAAAAA) {$\mytt{b(}\varphi_2\mytt{)} \vee \mytt{c(}\varphi_2\mytt{)} \leftarrow \mytt{a(}\varphi_2\mytt{)}$};
\node at (3,-4) [node](AAABAA) {$\mytt{b(}\varphi_2\mytt{)} \vee \mytt{c(}\varphi_2\mytt{)} \leftarrow \mytt{a(}\varphi_2\mytt{)}$};

\node at (-4.5,-4.8) [C_0](AAAAAAA) {$\mytt{d(}\varphi_2\mytt{)} \leftarrow \mytt{b(}\varphi_2\mytt{)}$};
\node at (-1.5,-4.8) [node](AAAAAAB) {$\mytt{d(}\varphi_2\mytt{)} \leftarrow \mytt{c(}\varphi_2\mytt{)}$};
\node at (1.5,-4.8) [node](AAABAAA) {$\mytt{d(}\varphi_2\mytt{)} \leftarrow \mytt{b(}\varphi_2\mytt{)}$};
\node at (4.5,-4.8) [node](AAABAAB) {$\mytt{d(}\varphi_2\mytt{)} \leftarrow \mytt{c(}\varphi_2\mytt{)}$};

\node at (-4.5,-5.6) [none](AAAAAAAA) {$\cdots$};
\node at (-3,-5.6) [none](AAAAAAAB) {$\cdots$};
\node at (-1.5,-5.6) [none](AAAAAABA) {$\cdots$};
\node at (0,-5.6) [none](AAAAAABB) {$\cdots$};
\node at (1.5,-5.6) [none](AAABAAAA) {$\cdots$};
\node at (3,-5.6) [none](AAABAAAB) {$\cdots$};
\node at (4.5,-5.6) [none](AAABAABA) {$\cdots$};
\node at (6,-5.6) [none](AAABAABB) {$\cdots$};

\path [line] (A) -- (AA);
\path [line] (A) -- (AB);
\path [line] (AA) -- (AAA);
\path [line] (AAA) -- (AAAA);
\path [line] (AAA) -- (AAAB);
\path [line] (AAAA) -- (AAAAA);
\path [line] (AAAA) -- (AAAAB);
\path [line] (AAAAA) -- (AAAAAA);
\path [line] (AAAAAA) -- (AAAAAAA);
\path [line] (AAAAAA) -- (AAAAAAB);
\path [line] (AAAB) -- (AAABA);
\path [line] (AAAB) -- (AAABB);
\path [line] (AAABA) -- (AAABAA);
\path [line] (AAABAA) -- (AAABAAA);
\path [line] (AAABAA) -- (AAABAAB);
\path [line] (AAAAAAA) -- (AAAAAAAA);
\path [line] (AAAAAAA) -- (AAAAAAAB);
\path [line] (AAAAAAB) -- (AAAAAABA);
\path [line] (AAAAAAB) -- (AAAAAABB);
\path [line] (AAABAAA) -- (AAABAAAA);
\path [line] (AAABAAA) -- (AAABAAAB);
\path [line] (AAABAAB) -- (AAABAABA);
\path [line] (AAABAAB) -- (AAABAABB);

\node[bigbox] [fit = (label) (A) (AAAAA) (AAABB)] {};

\end{tikzpicture}
\caption{The instantiation-tree for $P_{ex\mbox{-}\ref{ex:instantiation_tree}}$ and $\mytt{d(0)}$, and their stem.}\label{fig:instantiation_tree}
\end{figure}

\begin{example}\label{ex:instantiation_tree}
Consider a database atom $\mytt{d(0)}$ for the following program
$P_{ex\mbox{-}\ref{ex:instantiation_tree}}$:
\begin{alltt}\small
  \(\exists\)Y a(Y) \(\leftarrow\) d(X)             d(X) \(\leftarrow\) b(X)          e(X) \(\leftarrow\) d(X)
  b(X) v c(X) \(\leftarrow\) a(X)         d(X) \(\leftarrow\) c(X)
\end{alltt}\normalsize
The instantiation-tree is reported in Fig.~\ref{fig:instantiation_tree},
where we also highlighted the stem.
Note also that there are many isomorphic subtrees. This is due to a structural
property of $\funct{tree}(P \cup \cev{\at{a}})$, which we highlight in the
next lemma.
\qed
\end{example}

\begin{lemma}\label{lem:homo}
Let $m,n$ be two nodes of $T = \funct{tree}(P \cup \cev{\at{a}})$ s.t. $\body(\Label(m)) \simeq \body(\Label(n))$.
There is a node $m' \in T$ among $m$ and its siblings
s.t. $\funct{subtree}(m') \simeq \funct{subtree}(n)$.
\end{lemma}

\beforeProof\begin{proof}
The statement holds if $m = \Root(T)$ or $n = \Root(T)$ because in this case
$n = m$ as only the root of $T$ can contain a rule with an empty body.
Otherwise, let $m_p,n_p$ be the parent nodes of $m$ and $n$, respectively.
By construction (relying on Procedure~1),
$\body(\Label(m)) \subseteq \head(\Label(m_p))$ and
$\body(\Label(n)) \subseteq \head(\Label(n_p))$.
Let $\Label(n) = \hat{\sigma}(r)$, where $r$ is a rule and $\hat{\sigma}$ is a substitution.
Let $h$ be the isomorphism between $\body(\Label(m))$ and $\body(\Label(n))$.
Thus, there is a child $m'$ of $m$ s.t.
$\Label(m') = \widehat{h \circ \sigma}(r)$, which in turn implies
$\Label(m') \simeq \Label(n)$.
We now use induction.
Let $n,n_1,\ldots,n_k$ and $m',m_1',\ldots,m_k'$ ($k \geq 0$)
be two isomorphic paths in $\subtree(n)$ and $\subtree(m')$, respectively.
Still by construction,
there is a one-to-one mapping $\mu$ between the children of
$n_k$ and those of $m_k'$ s.t. for each child $n_{k+1}$ of $n_k$
it holds that $\Label(n_{k+1}) \simeq \Label(\mu(n_{k+1}))$.
\end{proof}

Given a model $M$ of $\funct{sinst}(P \cup \cev{\at{a}})$,
we shall show how to build a model $M^*$ of $\Inst(P \cup \cev{\at{a}})$
s.t. $M^* \models q$ implies $M \models q$.
%
%
Let $S = \funct{stem}(P \cup \cev{\at{a}})$, and $C_0$
be the smallest subset of $\nodes(S)$ satisfying the following properties:
(i) $\Root(S) \in C_0$;
(ii) $n \in C_0$ whenever its parent belongs to $C_0$ and
$\body(\Label(n)) \subseteq M$ holds.
We can thus restrict model $M$ as follows:
$M_0 = \{ \at{b} \in M \mid \at{b} \in \head(\Label(n)) \ \wedge \  n \in C_0 \}$.

\begin{example}
Consider again the instantiation-tree reported in
Fig.~\ref{fig:instantiation_tree}.
Let $M = \{\mytt{d(0)}\} \cup \{\mytt{a(}\varphi_i\mytt{)} \mid i \geq 1\}
\cup \{\mytt{b(}\varphi_i\mytt{)} \mid i \geq 1\}
\cup \{\mytt{c(}\varphi_i\mytt{)} \mid i \geq 2\}
\cup \{\mytt{d(}\varphi_i\mytt{)} \mid i \geq 1\}
\cup \{\mytt{e(}\varphi_i\mytt{)} \mid i \geq 1\}$.
Nodes in $C_0$ are those colored in gray, and
$M_0 = M \setminus \{\mytt{c(}\varphi_i\mytt{)} \mid i \geq 2\}$.
Note that $M_0$ is still a model of the program, as formally established
by the next lemma.
\qed
\end{example}

\begin{lemma}\label{lem:M0IsAModel}
If $M$ is a model of $\funct{sinst}(P \cup \cev{\at{a}})$, then also $M_0$ is.
\end{lemma}

\beforeProof\begin{proof}
Let $n \in \nodes(S) \setminus C_0$,
$m$ be the parent of $n$,
$\at{b}$ be the unique atom in $\body(\Label(n))$, and
$\at{b} \in M_0$.
We claim that $\head(\Label(n)) \cap  M_0 \neq \emptyset$.
By Procedure~1, since $\at{b} \in M_0$, then $\at{b} \in \head(\Label(m))$.
Moreover, according to the definition of $C_0$, if $m$ belongs to $C_0$, then also $n$ does.
Hence, $m \not\in C_0$ implying that there is a node $m'$ in $C_0$ s.t. $\at{b} \in \head(\Label(m'))$.
But this means, since $\at{b} \in M_0$, that there is a child $n'$ of $m'$ s.t. $n' \in C_0$
and $\Label(n') = \Label(n)$.
However, since by construction the head of each node in $C_0$ has a nonempty intersection with $M$,
then $\head(\Label(n')) = \head(\Label(n))$ has a nonempty intersection with $M_0$.
\end{proof}

From $T = \funct{tree}(P \cup \cev{\at{a}})$, we define
a total function $f : \nodes(T) \rightarrow \nodes(T)$ as follows:
For each node $n \in S = \funct{stem}(P \cup \cev{\at{a}})$, $f(n) = n$.
For the remaining nodes, let $n \in \nodes(T) \setminus \nodes(S)$ s.t.\ its
parent belongs to $S$. Let $m$ be the (unique) node in the path from
$\Root(T)$ to $n$ s.t.\ $\body(\Label(m)) \simeq \body(\Label(n))$.
Let $m'$ be either $m$ or one of its siblings according to whether
$\Label(m') \simeq \Label(n)$.
Function $f$ thus maps $\subtree(n)$ into $\subtree(m')$;
it is total by Lemma \ref{lem:homo}.
As a remark, we have that $n \simeq f(n)$, for each $n \in T$.
Moreover, $f(n) = n$ if and only if $n \in S$.

Finally, we build the set $C^*$ and the model
$M^*$ of $\Inst(P \cup \cev{\at{a}})$
s.t. $M^* \models q$ implies $M \models q$.
Initially, $C^*$ and $M^*$ coincide with $C_0$ and $M_0$, respectively.
Subsequently, for each node $n \in \nodes(T) \setminus \nodes(S)$ s.t.
both $\funct{parent}(n) \in C^*$ and $\body(\Label(n)) \subseteq M^*$,
$C^*$ is augmented by $n$ and $M^*$ is augmented by the set
$\{\at{b} \in \head(\Label(n)) \mid  h(\at{a}) \in M^*\}$
where $h$ is the isomorphism between $n$ and $f(n)$.

We now prove that QA can be performed by only
considering rules in the stem.

\begin{lemma}\label{lem:QAunderSinst}
It holds that $\Inst(P \cup \cev{\at{a}}) \models q$ if and only if $\funct{sinst}(P \cup \cev{\at{a}}) \models q$.
\end{lemma}

\beforeProof\begin{proof}
$(\Leftarrow)$
Since $\funct{sinst}(P \cup \cev{\at{a}}) \subseteq \Inst(P \cup \cev{\at{a}})$,
each model of $\Inst(P \cup \cev{\at{a}})$ is also a model of $\funct{sinst}(P \cup \cev{\at{a}})$.

\medskip
\noindent $(\Rightarrow)$
Let us assume that $\Inst(P \cup \cev{\at{a}}) \models q$ holds.
Let $M$ be a model of $\funct{sinst}(P \cup \cev{\at{a}})$.
Since, by construction,
$M^*$ is a model of $\Inst(P \cup \cev{\at{a}})$,
and since $M^* \models q$ by hypothesis, then $M \models q$ holds.
\end{proof}

Tractability of atomic QA over \lineardatalogexor can now be established.

\begin{theorem}\label{thm:logspace}
Data complexity of atomic QA over \lineardatalogexor programs is in \Logspace.
\end{theorem}

\beforeProof\begin{proof}
Armed with Lemma \ref{lem:QAunderSinst}, a logspace procedure iterates the database atoms looking for an atom $\at{a} \in D$
s.t. $\funct{sinst}(P \cup \cev{\at{a}}) \models q$.
In fact, for each $n \in \funct{stem}(P \cup \cev{\at{a}})$,
$\depth(n) < |\pi| \cdot (2w)^w$, where
$w$ is the maximum arity over all predicate symbols in $P$, and
$\pi$ is the number of predicate symbols occurring in $P$.
Therefore, cardinality of the ground program $\funct{sinst}(P \cup \cev{\at{a}})$
does not depend on $D$ and neither does the number of its minimal models,
which are sufficient for QA.
\end{proof}

\subsection{Discussion}





















\begin{table}[b]
\caption{Data complexity of QA in \datalogexor.}\label{tab:complexity}
\centering
\small\begin{tabular}{p{2.25cm}cccc}

  \cline{1-5}

  \multirow{2}[5]{2cm}{\textbf{\datalog Restrictions}} & \multirow{2}[5]{2cm}{\centering\textbf{Query Structure}} & \multicolumn{3}{|c}{\textbf{\datalog Extensions}} \bigstrut \\

  \cline{3-5}

  & & \multicolumn{1}{|c}{$\{\exists\}$} & $\{\vee\}$ & $\{\exists,\vee\}$ \bigstrut \\

  \cline{1-5}

  \multirow{2}[5]{2.6cm}{\vspace{.5em}\textsl{(Monadic-)}\linear} & AQ & \multicolumn{1}{|c}{in \ACzero} & in \Logspace & in \Logspace \bigstrut \\

                                    & ACQ/CQ & \multicolumn{1}{|c}{in \ACzero} & \coNPc & \coNPc \\

  \cline{1-5}

  \textsl{Multi-}\linear & AQ/ACQ/CQ & \multicolumn{1}{|c}{in \ACzero} & \coNPc & \coNPc \bigstrut \\

  \cline{1-5}

  \guarded & AQ/ACQ/CQ & \multicolumn{1}{|c}{\PTIMEc} & \coNPc & \coNPc \bigstrut \\

  \cline{1-5}

  \wguarded & AQ/ACQ/CQ & \multicolumn{1}{|c}{\EXPTIMEc} & \coNPc & \EXPTIMEc \bigstrut \\

  \cline{1-5}

\end{tabular}\normalsize
\end{table}

Table~\ref{tab:complexity} provides a comprehensive overview of complexity results
that follow from the results obtained in this section and in the
literature. Each row reports the complexity of QA for each of the classes defined in Section~\ref{sec:classes}
together with either atomic queries (AQ), acyclic conjunctive queries (ACQ) or
conjunctive queries (CQ).
In
each row we differentiate between the presence or absence of existential variables and disjunction:
$\exists$-variables in rule heads (column $\{\exists\}$), disjunctive
heads (column $\{\vee\}$), and both (column $\{\exists,\vee\}$).

Results in the $\{\exists\}$-column are from
\cite{CaliGottlobKiferKR2008,CaliGottlobLukasiewiczPODS09}, results
for \wguarded-\datalogor (last cell in column $\{\vee\}$) follow from
\citeText{EiterGottlobMannilaTODS1997}, since this class coincides with
\datalogor.
All the remaining \coNP-completeness results follow from Theorem \ref{thm:coNPcompl} in Section \ref{sec:complexGuard}, the remaining \EXPTIME-completeness results follow from Theorem \ref{thm:expcompl} in Section \ref{sec:complexWGuard}, and the \Logspace upper bounds follow from Theorem \ref{thm:logspace} in Section \ref{sec:complexAtomLin}.

Let us first consider the impact of allowing disjunction in the presence of
existential quantifiers in rule heads, i.e.\ columns $\{\exists\}$
versus $\{\exists,\vee\}$. We can see that in most considered cases,
the problem becomes (potentially) harder, except for the class
\wguarded. Indeed, for this case the problem is provably intractable
already without disjunctions, and turns out to remain so when
including them. In most other cases, we actually identify a
tractability boundary, passing from \ACzero to
\coNP-completeness. Notable exceptions are \textsl{Monadic-}\linear
and \linear with atomic queries, in which case the problem remains
tractable (but may be slightly more complex). It is interesting to
observe that in the presence of disjunction the nature of the query
has a huge impact on complexity for classes \textsl{Monadic-}\linear
and \linear, while this is not the case in the absence of disjunction.

Let us now discuss the impact of adding existential quantification
in the presence of disjunction in rule heads, i.e.\ columns
$\{\vee\}$ versus $\{\exists,\vee\}$. We can see that in all
considered classes except for \wguarded, adding existential
quantifiers does not alter complexity. This is a notable result, since
having existential quantification is a powerful construct for
knowledge representation. Only for \wguarded we obtain a
significant rise from \coNP-completeness to \EXPTIME-completeness and
thus provable intractability.

%
%
%
%
%
%
%
%
%



In future work, we intend to investigate on the exact data complexity
of atomic QA over (\textsl{Monadic-})\lineardatalogexor programs, in
particular whether it is in \ACzero{} or not. We also intend to study
the impact of disjunction on other tractable fragments of \datalogex
based on different paradigms, for example \emph{stickiness}
\cite{CaliGottlobPierisPVLDB10}, \emph{shyness} \cite{LeoneKR2012} and
\emph{weak-acyclicity}
\cite{FaginKolaitisMillerPopaTCS2005}. Moreover, it would also be
interesting to broaden the study to combined complexity or to
limit it to fixed or bounded predicate arities. Finally, also investigating
on implementation issues, for example in DLV$^\exists$ \cite{LeoneKR2012}, is on our
agenda.

\section{Acknowledgments}\label{ackn}
The authors want to thank Georg Gottlob, Michael Morak, and Andreas Pieris for useful discussions on the problem.
The work was partially supported by MIUR under the PON projects  FRAME
and TETRIS.

\end{document}